\documentclass{article}
\usepackage{ijcai21}

\usepackage{multirow}
\usepackage{booktabs}
\usepackage{amssymb}
\usepackage[ruled,linesnumbered]{algorithm2e}

\usepackage{color}

\usepackage{times}
\usepackage{soul}
\usepackage{url}
\usepackage[hidelinks]{hyperref}
\usepackage[utf8]{inputenc}
\usepackage[small]{caption}
\usepackage{graphicx}
\usepackage{amsmath}
\usepackage{booktabs}
\title{Dual Reweighting Domain Generalization for Face Presentation Attack Detection}

\author{
Shubao Liu$^{1,2}$\footnote{Equal Contribution}\and
Ke-Yue Zhang$^{2\ast}$\and
Taiping Yao$^{2\ast}$\footnote{Corresponding Author}\and
Kekai Sheng$^{2}$\and
Shouhong Ding$^{2\dag}$\and\\
Ying Tai$^{2}$\and
Jilin Li$^{2}$\and
Yuan Xie$^{1}$\And
Lizhuang Ma$^{1}$\\
\affiliations
$^1$East China Normal University, China\\
$^2$Youtu Lab, Tencent, Shanghai, China\\
\emails
shubaoL@stu.ecnu.edu.cn,
\{yxie, lzma\}@cs.ecnu.edu.cn,\\
\{zkyezhang, taipingyao, saulsheng, ericshding, yingtai, jerolinli\}@tencent.com
}

\begin{document}
% \linenumbers
\maketitle

\begin{abstract}
Face anti-spoofing approaches based on domain generalization (DG) have drawn growing attention due to their robustness for unseen scenarios. Previous methods treat each sample from multiple domains indiscriminately during the training process, and endeavor to extract a common feature space to improve the generalization. However, due to complex and biased data distribution, directly treating them equally will corrupt the generalization ability. To settle the issue, we propose a novel Dual Reweighting Domain Generalization (DRDG) framework which iteratively reweights the relative importance between samples to further improve the generalization.  Concretely, Sample Reweighting Module is first proposed to identify samples with relatively large domain bias, and reduce their impact on the overall optimization. Afterwards, Feature Reweighting Module is introduced to focus on these samples and extract more domain-irrelevant features via a self-distilling mechanism. Combined with the domain discriminator, the iteration of the two modules promotes the extraction of generalized features.
Extensive experiments and visualizations are presented to demonstrate the effectiveness and interpretability of our method against the state-of-the-art competitors.
\end{abstract}

%%%%%%%%% BODY TEXT

\begin{figure}[t!]
	\centering
	\includegraphics[width=1.0\linewidth]{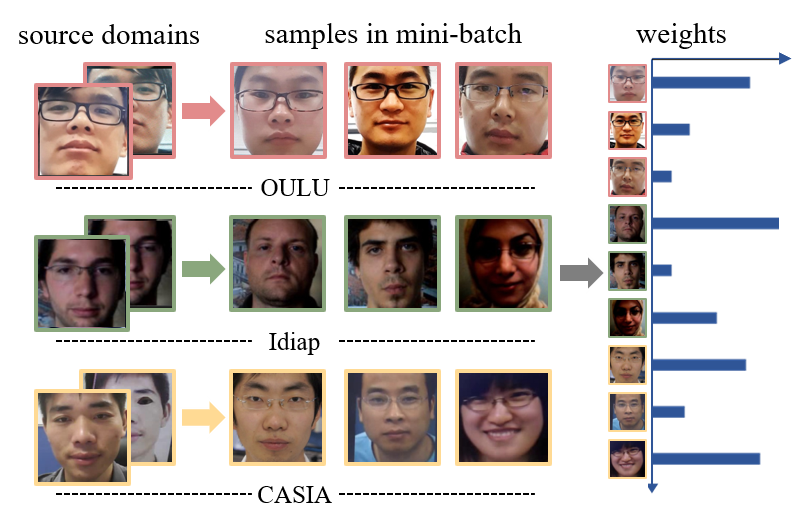}
	\vspace{-5mm}
	\caption{Previous face anti-spoofing methods treat each sample from multiple domains indiscriminately during the training process.  
	While our method iteratively reweights the relative importance between samples to further improve the generalization.}
	\vspace{-3mm}
	\label{illustration}
\end{figure}

\section{Introduction}
Face recognition techniques~\cite{Schroff2015FaceNetAU,Deng2019ArcFaceAA} have been widely utilized in our daily lives, such as smart device authentication, access control and entrance guard systems. Although bringing convenience, it also raises public concern about the safety issue since many kinds of face presentation attacks appear, such as photos and videos, which can easily fool the face recognition systems.
Therefore, many hand-crafted features based and deep learning based methods for presentation attack detection have been proposed in recent years. Hand-crafted methods utilize traditional feature descriptors, such as Local Binary Pattern (LBP)~\cite{LBP00}, Histogram of Oriented Gradients (HOG)~\cite{HoG00}, Scale Invariant Feature Transform (SIFT)~\cite{SIFT00}, \textsl{etc}. 
Standing in the strong representation abilities, Convolutional Neural Network (CNN) tackles this issue with different supervisions ~\cite{DeepBinary00,Zhang2021AuroraGR,additionInfo01}, such as binary cross-entropy, facial depth and remote-photoplethysmography (r-ppg) signal, \textsl{etc}.

Although the above state-of-the-art methods have attained remarkable performance in the intra-dataset setting where training and testing data come from the same distribution, the performance drops significantly encountering the testing data from different distributions. This is because the above methods don't take domain discrepancy into account, resulting in extracting biased features. 
To overcome the limitation, some researchers introduce the domain generalization (DG) methods for face presentation attack detection, which utilize several source domains to improve the generalization abilities of the model. Based on it, existing methods~\cite{shao2019multi,jia2020single} attempt to align the features of multiple source data equally for a shared feature space. However, for some samples are little domain-biased and some ones are large domain-biased, training on such samples indiscriminately will corrupt the generalization abilities of the methods.
Hence, how to alleviate such affection to learn a more generalizable representation remains a challenging problem.

To settle the problem, on one hand, inspired by curriculum learning, we argue that focusing on large domain-biased samples from the beginning will affect the overall optimization. However, on the other hand, solving these large domain-biased samples can further improve the generalization of the model. 
In order to stabilize the overall optimization direction and improve generalization, we propose a novel Dual Reweighting Domain Generalization (DRDG) framework for face anti-spoofing by iteratively reweighting the relative importance between samples, as illustrated in Figure~\ref{illustration}. Specifically, we propose two reweighting modules, Sample Reweighting Module (SRM) and  Feature Reweighting Module (FRM). Firstly, Sample Reweighting Module tries to identify samples with relatively large domain bias and reduce their impact on overall optimization. Secondly, Feature Reweighting Module (FRM) specially focuses on these large domain-biased samples and tries to extract more domain-irrelevant features via a self-distilling mechanism, which will promote generalization abilities. In general, these two modules are updated iteratively relying on the competition against the domain discriminator.
And the mutual iteration of these two modules ensures the overall optimization is not excessively affected by the large domain-biased data, while at the same time, further distills extracted features, and finally achieves better generalization.

The main contributions are summarized as follows:

$\bullet$ We propose a novel Dual Reweighting Domain Generalization framework for face anti-spoofing, which iteratively reweights the relative importance between samples from multiple source domains to further improve the generalization.  

$\bullet$ We specially design Sample Reweighting Module (SRM) and Feature Reweighting Module (FRM) to ensure that the overall optimization direction is not excessively affected by the large domain-biased data, and at the same time, pay more attention to distilling the feature of these data, resulting in the more generalizable model.

$\bullet$ Abundant experiments and visualizations are presented, which demonstrates the effectiveness of our method against the state-of-the-art competitors.

\section{Related Work}
\subsection{Face Anti-Spoofing}
In recent years, researchers have made great progress in the face anti-spoofing area. 
The development of it is concluded into two stages. Early researchers mainly utilized handcrafted feature descriptors, such as LBP~\cite{LBP00}, HOG~\cite{HoG00}, SIFT~\cite{SIFT00} and trained a traditional classifier for judgment. Some methods focused on temporal features to detect movement, such as eye blinking~\cite{eyeblink00}, head motion and lips motion.
With the rise of deep learning, ~\cite{DeepBinary00} regarded the face anti-spoofing as a binary classification task and leveraged the CNN architecture to solve it. 
To further boost the performance, ~\cite{additionInfo01} utilized additional supervisions, such as depth map, reflect map and r-ppg signal, to boost the performance. Based on auxiliary information, ~\cite{disentangle01,Liu2020OnDS} regularized features from the perspective of disentanglement. Since the above methods did not pay attention to the cross-dataset setting, MADDG~\cite{shao2019multi} utilized multiple domain discriminators to learn a generalized feature space. While SSDG~\cite{jia2020single} only aligned the features of real samples from different datasets but not for the features of the fake ones. Moreover, ~\cite{Chen2021GeneralizableRL} proposed $D^2AM$ to settle a more challenging generalizable scenario in the real world where domain labels are unknown.
However, these DG-based methods all attempt to align the samples straightforwardly and uniformly to a common feature space, regardless of the complexity and bias of data distribution, which leads to the degradation of generalization.

\begin{figure*}[t!]
	\centering
	\includegraphics[width=1.0\linewidth]{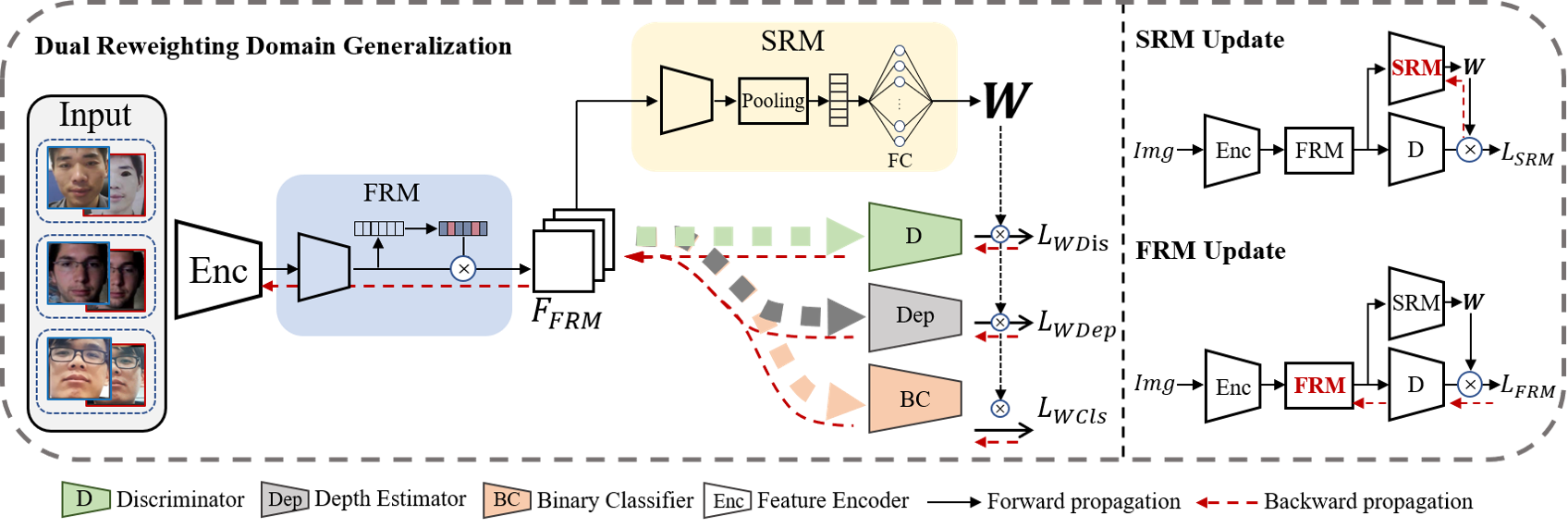}
	\caption{\textbf{The framework of our proposed method Dual Reweighting Domain Generalization (DRDG) for face presentation attack detection.}
	Besides the common modules, DRDG contains two novel modules named \textbf{Sample Reweighting Module (SRM)} and \textbf{Feature Reweighting Module (FRM)}, which reassign the most suitable weights to samples and features separately, leading to a discriminative and compact feature space. Moreover, we introduce the iterative optimization strategy, which further facilitates the cooperation of SRM and FRM for the better generalization on the unseen target domains.}
	\label{fig:framework}
\end{figure*}

\subsection{Domain Generalization}
Domain generalization aims to extract knowledge applicable to any unseen domain with multiple source domains, when target domain information is not available at training time. Several domain generalization methods have been proposed recently. ~\cite{ghifary2015domain} matched the feature distributions among source domains by using an auto-encoder. The work in ~\cite{li2018domain} based on adversarial learning to match the multiple domains to a pre-trained distribution. MLDG~\cite{li2018learning} proposed a model-agnostic meta-learning for domain generalization. \cite{Seo2020LearningTO} utilized batch and instance
normalizations to achieve the best combination for each domain. 
Based on the above DG methods, some face anti-spoofing methods have been proposed.
However, they directly treat these samples equally, which corrupts the generalization ability. While our framework, Dual Reweighting Domain Generalization (DRDG), iteratively reweights the relative importance between samples to further improve the generalization.

\section{Proposed Method}
\subsection{Overview}
Different from previous DG-based face anti-spoofing methods, which treat each sample from multiple domains indiscriminately during the training process, our proposed Dual Reweighting  Domain Generalization framework iteratively reweights the relative importance between samples to further improve the generalization.
Figure~\ref{fig:framework} illustrates the overview of our framework and the entire learning process. Specially, we design two reweighting modules, Sample Reweighting Module and Feature Reweighting Module, which are introduced in the Sec.~\ref{sec3.2} and Sec.~\ref{sec3.3}. 1) \textbf{Sample Reweighting Module} automatically reweights the relative importance of the samples in each mini-batch via measuring the feature transferability according to their domain information, and reduces the impact of large domain-biased samples for better overall optimization.
2) \textbf{Feature Reweighting Module} in turn focuses on these large domain biased samples, and attempts to extract more domain-irrelevant features via a self-distilling mechanism, which further promotes the generalization of the model. Combined with the iterative training process, these two modules facilitate the learning of generalized features for face presentation attack detection.

\subsection{Sample Reweighting Module}
\label{sec3.2}
Since some samples from multiple source domains contain large distribution discrepancies, aligning them at the early training stage inevitably introduces convergence problems and hinders the model’s generalization ability. To this end, we propose to dynamically adjust weights for different samples and reduce the impact of large domain-biased samples on the overall optimization. In our assumption, the samples, which are difficult to be distinguished for the domain discriminator, are the small domain-biased ones. On the contrary, the easily distinguished samples are the large domain-biased ones. Based on the assumption, we introduce a novel module Sample Reweighting Module (SRM) to assign weights for each sample via the adversarial procedure against the domain discriminator (described in Sec.~\ref{sec3.4}). Then, we utilize these weights to encourage the framework seeking a generalized feature space by bringing the distributions of high-weight samples much closer to each other and avoiding the influence of low-weight samples.

Particularly, SRM is responsible for assigning the weights to samples at each iteration, maximizing the loss of the domain discriminator. To keep up with the updated domain discriminator, SRM is forced to update constantly and reassign the weights across training. Considering the distribution discrepancies between the live and fake faces, we introduce two SRM modules, denoted as $SRM_{real}$ and $SRM_{fake}$, to reweight the samples separately for each class, which are denoted as SRM uniformly because of the same mechanism. SRM assigns weights to each sample $x_{i}$, which is denoted by $W_{i}$. Given a batch of samples, $x_{1}, x_{2},...x_{N}$, the number of samples in each domain is the same. SRM maps the features of these samples to an array of scores normalized via sigmoid function as the reweighting vector. During training, the optimization objective of SRM is formulated as Equation.\ref{SRM}, where $F_{FRM_i}$ is the feature map obtained from FRM, $D$ is Discriminator, $N$ is the number of one batch samples and $d_{i}$ is the domain label of image $x_i$, $W_i$ is the weight of image $x_i$, 
$\theta_{SRM}$ is the parameters of SRM.

\begin{equation}
    W_i = SRM(F_{FRM_i}; \theta_{SRM})
\end{equation}

\vspace{-2mm}

\begin{equation}
\label{SRM}
    \mathop{L_{SRM}} = \frac{1}{N}\sum^{N}_{i=1} W_i d_{i}log(D(F_{FRM_i}))
\end{equation}

\subsection{Feature Reweighting Module}
\label{sec3.3}
To make sure that our framework tackles the large domain-biased samples to facilitate learning a more generalizable representation finally, we propose the Feature Reweighting Module (FRM) through a self-distillation method to achieve this. Specifically, this module acts as an information modulator to self-distill common information from large domain-biased samples further, automatically selecting the proper feature channels to modulate contained information.
The modulated features amplify the domain-agnostic information and lessen the domain-relevant information.

Following the hypothesis in Sec.~\ref{sec3.2}, FRM employs it to self-distill the information of each sample via selecting the channels with different weights.
While interplaying with domain discriminator, FRM utilizes reverse weights $1-W_{i}$ to aggregate the large domain-biased samples much more for gradually tackling these samples, without impacting on other modules.
These two modules, FRM and SRM, facilitate each other to extract a robust representation, boosting the generalization performance. 
On one hand, FRM relies on the sample weights calculated from SRM. Hence FRM is updated after SRM in each iteration to fully utilize the domain-biased samples currently.
On the other hand, FRM suppresses the domain-relevant information of large domain-biased samples to help SRM out of the trap, improving the relative importance of such samples.

Concretely, the inputs of FRM are the features $F \in R^{N \times H \times W \times C}$ extracted by Feature Encoder, where $N$ is the number of images in each mini-batch, $H \times W$ is the spatial size of features and $C$ is the dimension of features. To ensure that each dimension is scaled with different scales, FRM calculates the weights for them via softmax function. During training, we minimize the reweighted loss of the domain discriminator $\mathcal{L}_{FRM}$ to obtain the weight $A_{i}$ as follows: 

\begin{equation}
    A_i = FRM(F_i; \theta_{FRM})
\end{equation}

\vspace{-2mm}

\begin{equation}
\label{FRM}
    \mathop{L_{FRM}} = \frac{1}{N}\sum^{N}_{i=1}(1-W_i)d_{i}log(D(A_i\otimes F_i))
\end{equation}
where $F_i$ is the feature map from Feature Encoder, $D$ is the  Discriminator, $N$ is the number of samples in one mini-batch and $d_{i}$ is the domain label of image $x_i$, $A_i$ is the channel-wise weight of feature $F_i$, 
$W_i$ is the weight of the sample $x_i$, $\theta_{FRM}$ is the parameters of FRM, $\otimes$ denotes the element-wise multiplication.

\subsection{Iterative Learning Framework}
\label{sec3.4}
\label{loss}
In this section, we briefly introduce other modules and the optimization strategy of our framework, as shown in Algorithm~\ref{optimization_strategy}.
Following the conventional DG-based methods for face anti-spoofing, we utilize Feature Encoder, Binary Classifier, Depth Estimator and Domain Discriminator in the framework. 
Concretely, \textbf{Feature Encoder} is proposed to extract domain-invariant features for better generalization on the unseen domains, which is optimized via $\mathcal{L}_{Enc}$.
\textbf{Binary Classifier} aims to detect the attacks from the real faces, which is optimized via $\mathcal{L}_{WCls}$.
\textbf{Depth Estimator} is introduced to estimate the facial depth map to guide the learning of the encoder. The pseudo-depth maps as grounding truth are produced by PRNet~\cite{PRNet}. Depth Estimator is optimized via $\mathcal{L}_{WDep}$.
\textbf{Domain Discriminator} is utilized to judge whether the extracted features are sufficiently domain-invariant, which is optimized via $\mathcal{L}_{Dis}$. 
And we only utilize the output of the classifier as the final results for face presentation attack detection.

To clearly explain the iterative learning process, here we describe one complete iterative unit in detail.

\noindent \textbf{Step A.}
We update the Discriminator (D) with the discriminator loss $\mathcal{L}_{Dis}$ firstly, because the two modules SRM and FRM compete against it, which requires the discriminator good enough to distinguish the domain of each sample. 

\noindent \textbf{Step B.}
Then SRM is utilized to estimate weights $W$ of each sample for focusing on the small domain-biased ones. Combining $W$ and the discriminator loss $\mathcal{L}_{Dis}$, we update SRM to minimize the loss $\mathcal{L}_{SRM}$, which improves the estimation abilities of SRM for more accurate reweighting.

\noindent \textbf{Step C.}
Next step, the weights $W$ are utilized to calculate other loss functions, $\mathcal{L}_{WDep}$, $\mathcal{L}_{WCls}$ and $\mathcal{L}_{Enc}$. We update Depth Estimator (Dep), Binary Classifier (BC), Feature Encoder (Enc) and (FRM) with the above loss.

\noindent \textbf{Step D.}
Repeating the above steps K-1 times, we utilize FRM to handle the large domain-biased samples and extract more generalizable representation further via the self-distillation mechanism. Then, we update FRM only by the loss $\mathcal{L}_{FRM}$, which is calculated by the discriminator loss $\mathcal{L}_{Dis}$ and the reverse weights $1-W$ of the corresponding iteration.

\begin{algorithm}[t!]
    \caption{The optimization strategy of our DRDG}
    \label{optimization_strategy}
    \KwData{$x_b^{i}$: Batch size for samples from domain $i$}
    Initial model parameters and determine N, K, $\lambda_{1}$, $\lambda_{2}$\;
    Shuffle all samples from different domains\;
    \For{$t$ in (1:N)} {
        \uIf {t==1}{
            Initial $W$ = 1\;
        }
        \uIf {t \% K!=0}{
            \uIf {t \% 2!=0}{
                Calculate $\mathcal{L}_{Dis}= \frac{1}{N}\sum^{N}_{i=1}- d_{i}log(D(F_{FRM_i}))$\;
                Update $\theta_{D}$ with $\mathcal{L}_{Dis}$\;    
            }
            \uElse{
                Calculate $\mathcal{L}_{SRM}$ as Eq.~\ref{SRM}\;
                Update $\theta_{SRM}$ with $\mathcal{L}_{SRM}$ and update $W$\;
                Calculate $\mathcal{L}_{WDep}= \frac{1}{N}\sum^{N}_{i=1} W_i \left\| Dep(F_{FRM_i}) - I_i \right\|_{2}^{2}$\;
                Calculate $\mathcal{L}_{WCls}=\frac{1}{N}\sum^{N}_{i=1} -W_i y_i log(BC(F_{FRM_i}))$\;
                Update $\theta_{Dep}$, $\theta_{BC}$ separately\;
                Calculate $\mathcal{L}_{Enc}=\mathcal{L}_{WCls}+\lambda_{1}\mathcal{L}_{WDep}-\lambda_{2}W\mathcal{L}_{Dis}$\;
                Update $\theta_{Enc}$ and $\theta_{FRM}$ with $\mathcal{L}_{Enc}$\;
            }
        }
        \uElse{
            Calculate $\mathcal{L}_{FRM}$ as Eq.~\ref{FRM}\;
            Update $\theta_{FRM}$ with $\mathcal{L}_{FRM}$\;
        }
    }
    \Return Model parameters\;
\end{algorithm}

\begin{figure*}[t!]
	\centering
	\setlength{\abovecaptionskip}{0.1cm}
	\includegraphics[width=1.0\linewidth]{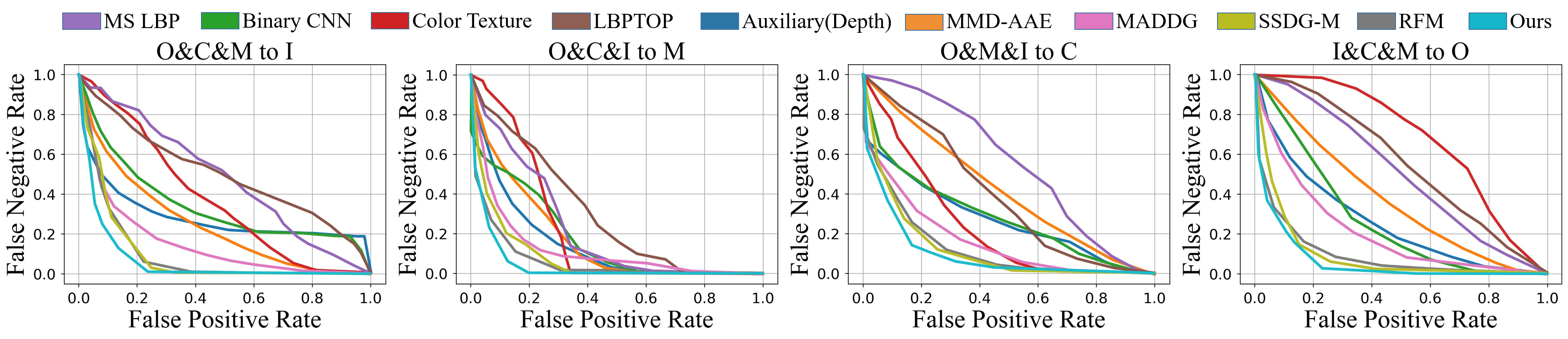}
	\caption{ROC curves of four testing tasks for domain generalization on face anti-spoofing.}
	\label{fig:ROC}
\end{figure*}

\begin{table*}[t!]
    \begin{center}
	\scalebox{0.8}{
    \begin{tabular}{ccccccccc}
   \toprule
   \multirow{2}{*}{\textbf{Method}}&
    \multicolumn{2}{c}{\textbf{O\&C\&M to I}}&\multicolumn{2}{c}{\textbf{O\&C\&I to M}}&\multicolumn{2}{c}{\textbf{O\&M\&I to C}}&\multicolumn{2}{c}{\textbf{I\&C\&M to O}}\cr
    \cmidrule(lr){2-3} \cmidrule(lr){4-5} \cmidrule(lr){6-7} \cmidrule(lr){8-9}
    &HTER(\%)&AUC(\%)&HTER(\%)&AUC(\%)&HTER(\%)&AUC(\%)&HTER(\%)&AUC(\%)\cr
    \midrule
    MS\_LBP \cite{LBP03} &50.30&51.64&29.76&78.50&54.28&44.98&50.29&49.31\cr
    Binary CNN \cite{DeepBinary03} &34.47&65.88&29.25&82.87&34.88&71.94&29.61&77.54\cr
    IDA \cite{wen2015face} &28.35&78.25&66.67&27.86&55.17&39.05&54.20&44.59 \cr
    Color Texture \cite{other01} &40.40&62.78&28.09&78.47&30.58&76.89&63.59&32.71 \cr
    LBPTOP \cite{de2014face} &49.45&49.54&36.90&70.80&42.60&61.05&53.15&44.09 \cr
    Auxiliary(Depth Only)&29.14&71.69&22.72&85.88&33.52&73.15&30.17&77.61 \cr
    Auxiliary(All) \cite{additionInfo01} &27.60&-&-&-&28.40&-&-&-\cr
    MMD-AAE \cite{li2018domain} &31.58&75.18&27.08&83.19&44.59&58.29&40.98&63.08 \cr
    MADDG \cite{shao2019multi} &22.19&84.99&17.69&88.06&24.50&84.51&27.98&80.02 \cr
    SSDG-M \cite{jia2020single} &18.21&\textbf{94.61}&16.67&90.47&23.11&85.45&25.17&81.83\cr
    RFM \cite{shao2020regularized} &17.30&90.48&13.89&93.98&20.27&88.16&16.45&91.16 \cr
    \midrule
    \textbf{DRDG (Ours)}&\textbf{15.56}& 91.79 &\textbf{12.43}&\textbf{95.81}&\textbf{19.05}&\textbf{88.79}&\textbf{15.63}&\textbf{91.75}\cr
    \bottomrule
    \end{tabular}
    }
    \end{center}
    \vspace{-4mm}
    \caption{Comparison results on four testing tasks for domain generalization on face anti-spoofing.}
    \label{tab:anti_1}
    \vspace{-2mm}
\end{table*}

\section{Experiments}
\subsection{Experimental Settings}

\textbf{Datasets.} We strictly follow the previous works~\cite{shao2019multi,shao2020regularized,jia2020single}, utilizing four public databases, to evaluate our method: OULU-NPU (denoted as O)~\cite{Oulu}, CASIA-FASD (denoted as C)~\cite{CASIA}, MSU-MFSD (denoted as M)~\cite{wen2015face} and Idiap Replay-Attack (denoted as I)~\cite{Replay}.
Concretely, we select one dataset for testing and the remaining three for training, then four testing tasks are obtained: O\&C\&M to I, O\&C\&I to M, O\&M\&I to C and I\&C\&M to O.
Since significant
domain shift exists among these datasets, \textit{e.g.,} materials, illumination, background, resolution and \textit{etc.}, the tasks are challenging.

\noindent \textbf{Implementation Details.} Our method is implemented via PyTorch on $11G$ NVIDIA $2080$Ti GPUs with Linux OS and trained with Adam optimizer~\cite{kingma2014adam}. The learning rates $\alpha$, $\beta$ are set as 1e-3, 1e-4, respectively. We extract RGB and HSV channels of images, thus the input size is $256\times256\times6$. In training phase, the balance coefficients $\lambda_{1}$ and $\lambda_{2}$ are set to $10$ and $0.1$ respectively. In our method, the K in Algorithm \ref{optimization_strategy} is an important hyperparameter, which determines the training pace of large domain-biased samples, and we set it as 5 according to our experiments. We strictly follow the popular evaluation metrics,  which contain the Half Total Error Rate (HTER) and the Area Under Curve (AUC).

\subsection{Comparison Results}
As shown in Table~\ref{tab:anti_1} and Figure~\ref{fig:ROC}, our method outperforms all state-of-the-art methods. Firstly, we can observe that the DG-based face anti-spoofing methods perform better than conventional methods. This proves that the distribution of target domain is different from source domain, while conventional methods only focus on the spoofing cues fitting source domain. Moreover, the proposed method outperforms other DG-based methods.
Concretely, compared to MADDG~\cite{shao2020regularized}, DRDG gains $5.26\%$ and $5.45\%$ improvement on the task O\&C\&I to M and O\&M\&I to C respectively with HTER.
We believe the reason is that our method focuses on small domain-biased samples and self-distill the common information from large domain biased ones, the cooperation of which facilitate the generalization. 
As shown in Table~\ref{tab:anti_2}, we also evaluate our method with extremely limited source domains, and it achieves the best performance, which verifies its effectiveness again.
As for computational efficiency, since SRM is discarded and only FRM is retained during the inference stage, our method is comparable to others.

\begin{table}
\footnotesize
\begin{center}
\scalebox{0.85}{
\begin{tabular}{ccccc}
   \toprule
   \multirow{2}{*}{\textbf{Method}}&
    \multicolumn{2}{c}{\textbf{M\&I to C}}&\multicolumn{2}{c}{\textbf{M\&I to O}}\cr
    \cmidrule(lr){2-3} \cmidrule(lr){4-5}
    &HTER(\%)&AUC(\%)&HTER(\%)&AUC(\%)\cr
    \midrule
    MS\_LBP &51.16&52.09&43.63&58.07\cr
    IDA&45.16&58.80&54.52&42.17\cr
    Color Texture&55.17&46.89&53.31&45.16 \cr
    LBPTOP&45.27&54.88&47.26&50.21 \cr
    MADDG&41.02&64.33&39.35&65.10 \cr
    SSDG-M&31.89&71.29&36.01&66.88 \cr
    \midrule
    \textbf{DRDG (Ours)}&\textbf{31.28}&\textbf{71.50}&\textbf{33.35}&\textbf{69.14} \cr
    \bottomrule
    \end{tabular}
    }
    \end{center}
    \vspace{-3mm}
    \caption{Comparison results on limited source domains.}
    \label{tab:anti_2}
    \vspace{-3mm}
\end{table}

\begin{table}
    \footnotesize
    \begin{center}		
    % \scalebox{1.0}{
    \setlength{\tabcolsep}{4mm}{
    \begin{tabular}{ccc}
   \toprule
   \textbf{Method}&HTER(\%)&AUC(\%)\cr
    \midrule
    Baseline&17.86&90.93 \cr
    Baseline\_FRM&15.31&91.21 \cr
    Baseline\_FRM\_reverse&19.34&89.82 \cr
    Baseline\_SRM&13.57&94.05 \cr
    Baseline\_SRM\_reverse&20.67&88.15 \cr
    \textbf{DRDG}&\textbf{12.43}&\textbf{95.81} \cr
    \bottomrule
    \end{tabular}}
    \end{center}
    \vspace{-3mm}
    \caption{Evaluations of different components of the proposed method on \textbf{O\&C\&I to M} task.}
    \label{tab:anti_3}
    \vspace{-3mm}
\end{table}

\begin{figure*}[t!]
	\centering
	\setlength{\abovecaptionskip}{0.2cm}
	\scalebox{1.0}{
    	\includegraphics[width=1.0\linewidth]{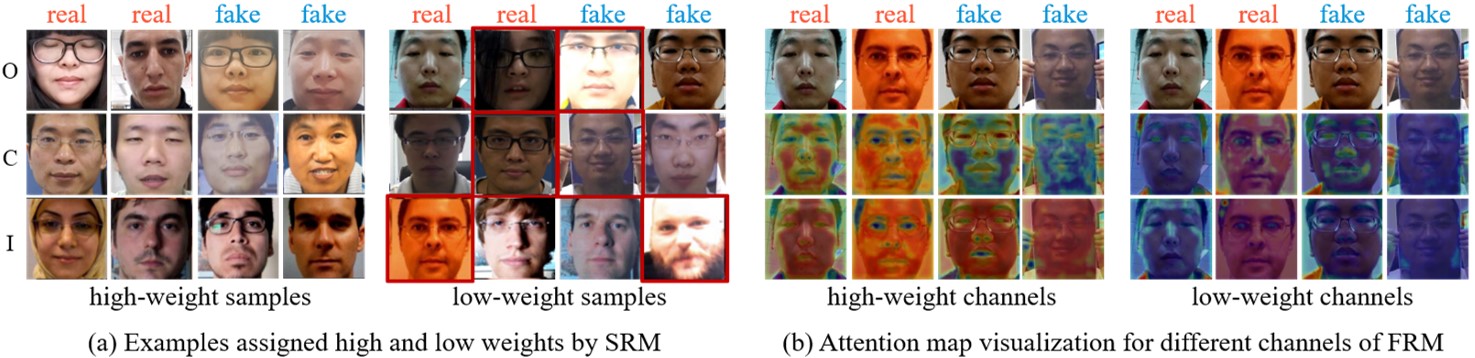}}
	\caption{The demonstration of high-weight and low-weight samples assigned by SRM and attention maps for high-weight and low-weight channels of FRM. The samples and attention maps are collected according to a model randomly selected at a certain epoch during training.}
	\label{fig:CAM}
	\vspace{-2mm}
\end{figure*}

\subsection{Ablation Study}
We perform the ablation study to verify the contribution of each component on O\&C\&I to M task. As shown in Table~\ref{tab:anti_3}, several observations are made as follows: 
1) Compared with Baseline where both SRM and FRM are removed from DRDG, Baseline\_FRM gains a better performance, which demonstrates that FRM enhances the features via a self-distillation mechanism to extract a more common feature for generalization. 
Moreover, Baseline\_FRM\_reverse aims to push the samples from different domains apart, which performs even worse than Baseline. 
2) Baseline\_SRM performs better than Baseline, while Baseline\_SRM\_reverse which emphasizes large domain-biased samples first and small domain-biased ones later performs worse than it. 
It proves the correctness of our training strategy via starting from the small domain-biased ones than dealing with the large domain-biased ones.
3) DRDG achieves the best performance compared to the other variants, which confirms that FRM and SRM facilitate each other to extract a robust representation and boost the generalization performance. 

\begin{figure}[t!]
	\centering
	\setlength{\abovecaptionskip}{0.2cm}
	\scalebox{1.0}{
    	\includegraphics[width=1.0\linewidth]{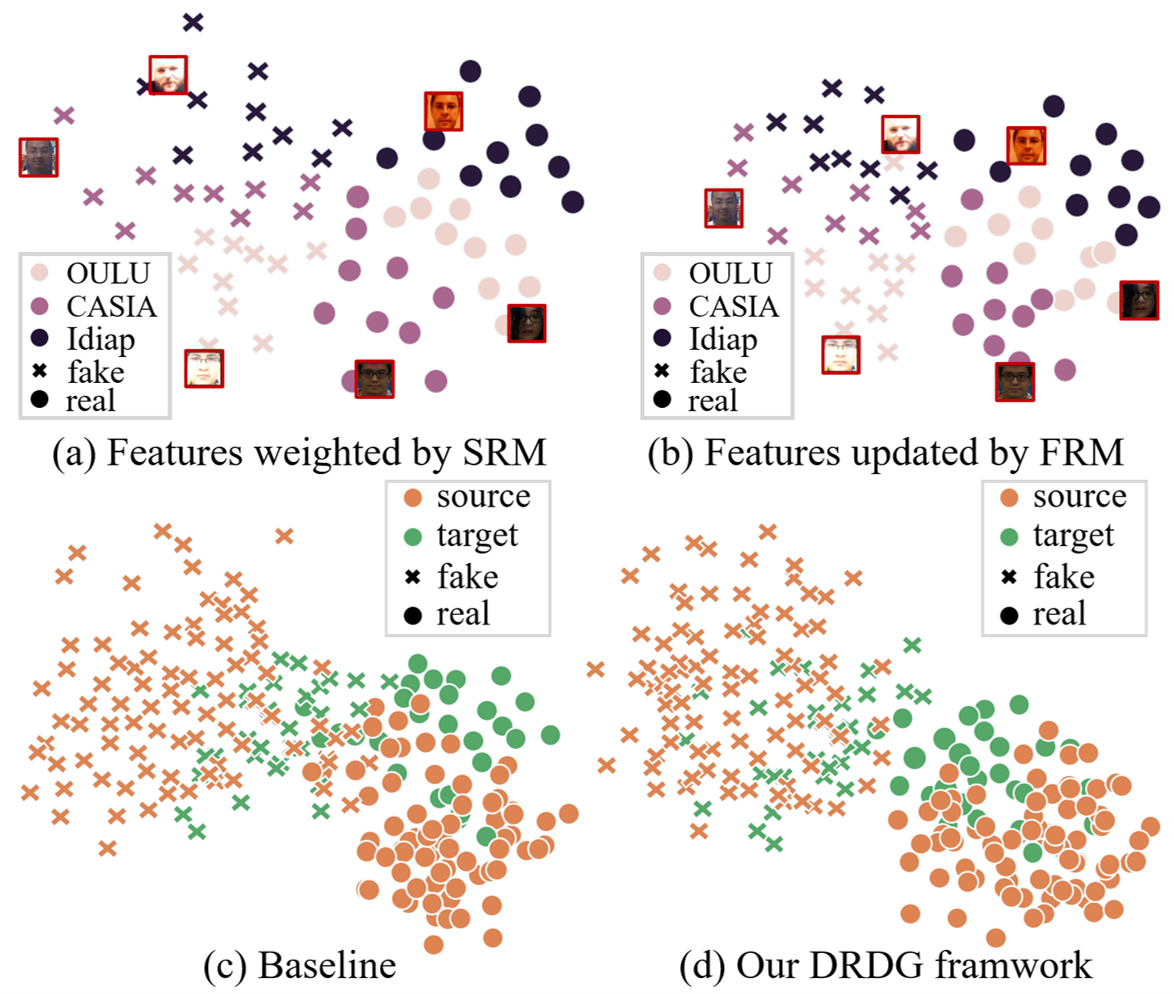}}
	\caption{Feature distributions of different stages and models.}
	\label{fig:tSNE-weight}
	\vspace{-3mm}
\end{figure}

\subsection{Visualization and Analysis}

\textbf{Sample Reweighting Visualization.}
To verify the effectiveness of SRM, we randomly select a model at a certain epoch during training for visualization, which is conducted on O\&C\&I to M task.
We first demonstrate some high-weight and low-weight samples assigned by SRM in three source domains, which are from the same mini-batch, as shown in Figure~\ref{fig:CAM}(a).
It is obvious that the differences between low-weight samples are significantly larger than those between high-weight ones. It is attributed to their large distribution discrepancies, \textit{e.g.}, sample with overexposed light and one with reddish overall style (all marked with red boxes). 
Focusing on them may mislead the optimization of the network, hence SRM assigns the lower weights to them, alleviating their impact.
As shown in Figure~\ref{fig:tSNE-weight}(a), We also visualize the feature distributions of these samples via t-SNE.
We find that high-weight samples aggregate much more closer to each other and low-weight samples are much far away from the center like outliers.  Aligning these samples equally corrupts the generalization ability of our framework.
Hence, SRM indeed ensures that our framework automatically selects the most suitable samples for alignment in each mini-batch, facilitating the overall optimization for better generalization.

\noindent\textbf{Feature Reweighting Visualization.}
Following the experiments of SRM, we utilize the same model to draw the visualization for FRM.
To make an exploration about the channels with different weights, we demonstrate the feature maps of the low-weight samples selected from SRM, as illustrated in Figure~\ref{fig:CAM}(b).
We find that the attention maps of low-weight channels are with low energy and focus on some trivial cues, such as backgrounds, clothes, \textit{etc}, which are not generalizable.
While attention maps of high-weight channels are always with high energy and mainly pay attention to the region of the internal face, which are more likely to be intrinsic and generalized. 
Hence the self-distillation mechanism of FRM indeed promotes the extraction of common features for better generalization abilities.
The distribution of the features extracted by the updated FRM is also illustrated in Figure~\ref{fig:tSNE-weight}(b). 
Compared with Figure~\ref{fig:tSNE-weight}(a), we find that the large domain-biased samples are pulled back to the center after the update of FRM, which verifies that FRM tackles such samples to get a more compact feature space.

\noindent\textbf{Feature Distribution.}
To study the effect of the framework containing two modules on the feature distributions, we draw the visualization of Baseline and framework DRDG, as shown in Figure~\ref{fig:tSNE-weight}(c),(d) separately.
Since baseline tries to align all the samples regardless of their distribution discrepancies, the samples relatively scatter in the feature space, which results in the performance degradation on the target domain.
While feature space constructed by framework DRDG is more compact than Baseline overall, it achieves better class boundary and generalizes well on target domain for two modules facilitating each other.

\section{Conclusion}
In this paper, we propose a new perspective of FAS that automatically selects the most suitable samples for domain generalization. 
Concretely, we present an iteratively learning framework, Dual Reweighting Domain Generalization, containing two novel modules, to boost the generalization abilities. 
SRM and FRM facilitate each other to establish the training strategy via reassigning the most suitable weights to samples and features respectively, and pull all the samples aggregated for better generalization results.
Comprehensive experiments and visualizations are presented to demonstrate the effectiveness and interpretability of our method against the state-of-the-art competitors.

\clearpage
\small{
\bibliographystyle{named}
\bibliography{ijcai21}
}
\end{document}